\documentclass{article}

\usepackage{amsmath,amsfonts}
\usepackage{amssymb}
\usepackage{float}
\usepackage{graphicx}
 \usepackage{verbatim}

\title{HyperFitS - Hypernetwork Fitting Spectra for metabolic quantification of ${}^1$H MR spectroscopic imaging}

\author{
Paul J.~Weiser$^{1,2,3}$,
Gülnur~Ungan$^{1,2}$, \\
Amirmohammad~Shamaei$^{4,5}$, 
Georg~Langs$^{3}$, \\
Wolfgang~Bogner$^{6}$,
Malte~Hoffmann$^{1,2}$, \\
Antoine~Klauser$^{7}$,
Ovidiu~C.~Andronesi$^{1,2}$
}

\begin{document}


\maketitle

\begin{center}
\small
$^{1}$ Athinoula A. Martinos Center for Biomedical Imaging, Massachusetts General Hospital, Boston, MA, USA\\
$^{2}$ Department of Radiology, Massachusetts General Hospital, Harvard Medical School, Boston, MA, USA\\
$^{3}$ Computational Imaging Research Lab – Department of Biomedical Imaging and Image-Guided Therapy, 
Medical University of Vienna, Vienna, Austria\\
$^{4}$ Electrical and Software Engineering, University of Calgary, Calgary, Canada\\
$^{5}$ Hotchkiss Brain Institute, University of Calgary, Calgary, Canada\\
$^{6}$ High Field MR Center – Department of Biomedical Imaging and Image-Guided Therapy, 
Medical University of Vienna, Vienna, Austria\\
$^{7}$ Advanced Clinical Imaging Technology, Siemens Healthineers International AG, Lausanne, Switzerland
\end{center}
\date{}

%
%
%
%

\begin{abstract}
\noindent
\textbf{Purpose.} Proton magnetic resonance spectroscopic imaging ($^1$H MRSI) enables the mapping of whole-brain metabolites concentrations in-vivo. However, a long-standing problem for its clinical applicability is the metabolic quantification, which can require extensive time for spectral fitting. Recently, deep learning methods have been able to provide whole-brain metabolic quantification in only a few seconds. However, neural network implementations often lack configurability and require retraining to change predefined parameter settings. \newline
\textbf{Methods.} We introduce HyperFitS, a hypernetwork for spectral fitting for metabolite quantification in whole-brain $^1$H MRSI that flexibly adapts to a broad range of baseline corrections and water suppression factors. Metabolite maps of human subjects acquired at 3T and 7T with isotropic resolutions of 10 mm, 3.4 mm and 2 mm by water-suppressed and water-unsuppressed MRSI were quantified with HyperFitS and compared to conventional LCModel fitting.  \newline
\textbf{Results.} Metabolic maps show a substantial agreement between the new and gold-standard methods, with significantly faster fitting times by HyperFitS. Quantitative results further highlight the impact of baseline parametrization on metabolic quantification, which can alter results by up to 30\%. \newline
\textbf{Conclusion.} HyperFitS shows strong agreement with state-of-the-art conventional methods, while reducing processing times from hours to a few seconds. Compared to prior deep learning based spectral fitting methods, HyperFitS enables a wide range of configurability and can adapt to data quality acquired with multiple protocols and field strengths without retraining. \newline

\end{abstract}

\section{Introduction}

    \begin{figure*}[ht]
		\centering
		\includegraphics[width=1.0\linewidth]{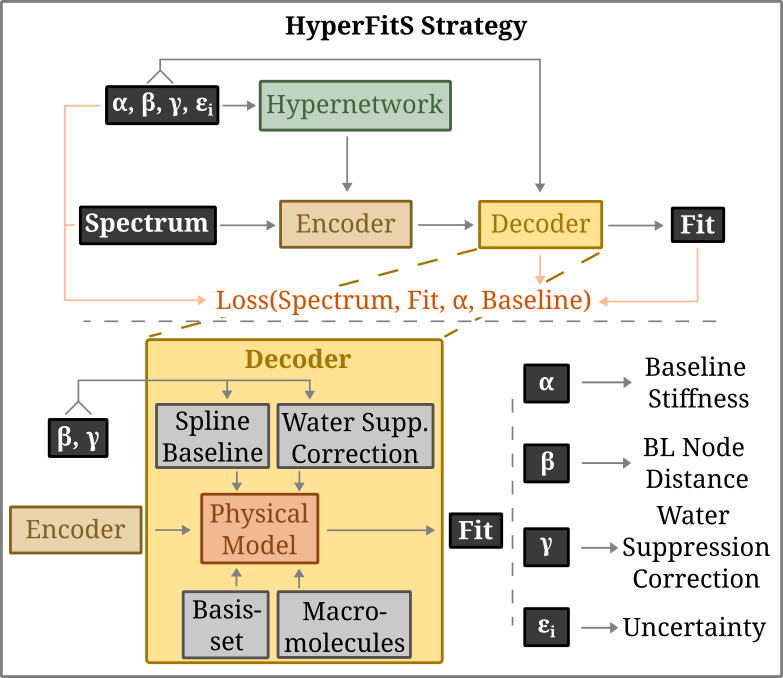}
		\caption{HyperFitS: Top: The HyperFitS strategy showing a hypernetwork taking baseline and water suppression correction parameters and predicting weights for a quantification network with a physics-based encoder. Bottom: The physical model combining metabolite and macromolecule basis set, baseline spline functions, and water suppression correction factor used to predict the spectral fit.}
        \label{fig:overview}
	\end{figure*}
    
	Magnetic Resonance Spectroscopic Imaging (MRSI) enables mapping of metabolite concentrations in-vivo. For instance, 3-dimensional whole-brain MRSI allows for the quantification of up to 20 individual metabolites \cite{klauser2024eccentric, hangel2020high, lam2016high}, which can help identify and diagnose metabolic abnormalities that occur in glioblastoma or neurodegenerative diseases \cite{hangel2020high}. However, the interpretation of MRSI data for clinical applications or basic research hinges on accurate quantification of individual metabolite concentrations in each spectrum, which requires spectral fitting \cite{wilson2019methodological, provencher1993estimation, near2021preprocessing}.
    
    There are multiple fitting software packages that enable in-vivo spectral quantification \cite{provencher1993estimation, ratiney2005time, wilson2011constrained, clarke2021fsl, soher2023vespa, chong2011two, juchem2016inspector, naressi2001java, maudsley2009mapping, maudsley2006comprehensive, crane2013sivic}. Out of these, LCModel \cite{provencher1993estimation} has become a gold standard for single voxel and 3D ${}^1$H MRSI. LCModel performs spectral fitting by linearly combining metabolite spectra from a basis set and estimates the concentration of individual metabolites by the weighting coefficient for the area under curve of the corresponding metabolite spectrum in the basis set. 

    Although LCModel employs an elaborate signal model, it does not capture the entire complexity associated with all the technical aspects involved in the data acquisition. The accurate quantification of metabolite concentrations can be impacted by artifacts that distort the spectral signal. In particular, distortion of the baseline can have a large effect on the accuracy and stability of quantification models \cite{near2021preprocessing, cudalbu2021contribution}. Imperfections such as insufficient water suppression during acquisition can complicate the baseline and introduce large signal variability. During spectral fitting, a baseline is modeled using constrained splines and most methods adjust the spline amplitude only for a fixed number of knots. Optimal parametrization for baseline modeling requires adjusting both degrees of freedom, which is challenging with approaches such as LCModel. On the other hand, water suppression may reduce the amplitude of metabolite peaks close to the water peak which needs to be corrected during metabolite quantification. LCModel provides correction only for a single metabolite (creatine), however the transition band of water suppression pulses affects multiple metabolites. All these spectral artifacts and correction methods contribute to a level of uncertainty during the spectral quantification which is often approximated by the lower bound of the variance computed in the form of the Cramer-Rao lower bounds (CRLBs). 
    
    A major drawback of conventional methods such as LCModel are their long processing times for MRSI data. While parallelization can speed-up fitting of multi-voxel data, the processing time for a single 3D MRSI at 3.4mm resoltion can take up 1h \cite{weiser2025deep}. Therefore, deep learning methods have been developed in recent years, which allow for GPU utilization and near instantaneous quantification of whole-brain MRSI volumes in only a few seconds \cite{gurbani2019incorporation}.
    
    In contrast to conventional methods, which fit each spectrum individually in an iterative manner, deep learning-based models learn optimal spectral quantification during prior extensive training and provide fast results during inference \cite{shamaei2023physics, gurbani2019incorporation, osburg2025deep, turco2024tensorfit, hiltunen2002quantification, bhat2006fast, hatami2018magnetic, lee2019intact, chandler2019mrsnet, shamaei2021wavelet, rizzo2023quantification}. 
    Popular training paradigms for deep learning based metabolite quantification include supervised and self-supervised optimization \cite{merkofer2025strategies}. In the supervised case, a model is trained to directly predict metabolite levels from the input spectrum and minimize a loss between the predictions and ground-truth concentrations without predicting a fitted spectrum \cite{hatami2018magnetic, lee2019intact, chandler2019mrsnet, shamaei2021wavelet, iqbal2021deep}. In the self-supervised case, the network uses a physical model that includes spectra of individual metabolites and predicts a fitted spectrum as a combination of individual spectra. The measured spectrum is used as a training target and the network predicts parameters of the physical model, which are modeled into the fitted spectrum. The loss is derived from the difference between the input and the modeled spectrum \cite{shamaei2023physics, gurbani2019incorporation, chen2024magnetic}. Latest developments include uncertainty estimation and the possibility of providing baseline optimization from a discrete set of predefined constraints \cite{shamaei2026phive}. 

    However, while deep learning provides significantly reduced processing times, previously developed quantification models have some limitations: they 1) lack water suppression correction, 2) only allow a single or discrete number of baseline constraints, and 3) do not provide uncertainty estimation or require the training of multiple separate models as a surrogate for uncertainty estimation. Eliminating these limitations would allow the fitting to adapt to diverse spectral data quality stemming from different acquisition protocols and field strengths.
    
    To extend the performance of deep learning spectral fitting models we adopt a hypernetwork strategy \cite{schmidhuber1993self, ha2016hypernetworks, hoopes2021hypermorph}. In this work, we present HyperFitS, a hypernetwork for the quantification of whole-brain MRSI at 3T and 7T. The hypernetwork introduces a binary input parameter to switch water suppression correction on or off, two continuous parameters controlling baseline flexibility and spline node distance during inference. Additionally, our approach provides aleatoric and epistemic uncertainty estimations  with a single trained hypernetwork instead of training multiple separate quantification models \cite{krueger2017bayesian}.

\section{Methods}
    
    \subsection{Strategy}
        Spectral quantification methods linearly combine a metabolic basis set to provide a fit $f$ of a given spectrum $s'$. Here, an unsupervised deep learning approach is utilized: An encoder $\mathcal{E}$ receives an in-vivo spectrum $s'$ as input, and predicts parameters $p$ for the decoder $\mathcal{D}$. During training, a loss is minimized between the predicted fit $f$ and the phase corrected target spectrum $s$.
        \begin{align}
            \mathcal{L}(\mathcal{D}\circ\mathcal{E}(s'),s)
        \end{align}
        We extended this approach by proposing a hypernetwork $\mathcal{H}$ that takes hyperparameters $\theta$ as input and predicts a set of weights $\mathcal{H}(\theta)=\Omega$ for the encoder $\mathcal{E}(.|\Omega)$. The weights of the encoder $\mathcal{E}$ are determined by the hypernetwork. All learnable weights are located in the hypernetwork.
        \begin{align}
            \mathcal{L}(\mathcal{D}\circ\mathcal{E}(s'|\mathcal{H}(\theta)),s)
        \end{align}
        The hypernetwork receives a total of 3+I hyperparameters $\theta = \{\alpha, \beta, \gamma, \epsilon_1,...,\epsilon_I\}$ as input.
        The overall strategy is illustrated by the block diagram in Figure \ref{fig:overview}. Each hyperparameter affects a different functionality of the spectral quantification model:
        \begin{itemize}
            \item \textbf{Baseline ($\alpha,\beta$):} The baseline is modeled by spline functions that interpolate between spline nodes. Hyperparameters $\alpha$ and $\beta$ determine the baseline flexibility and spline node distance, respectively. (Section \ref{sec:baseline})
            \item \textbf{Water suppression correction ($\gamma$):} Water suppression artifacts are corrected by multiplying the fit $f$ with a learnable inverted Gaussian function $\omega$. Water suppression correction is enabled by the binary parameter $\gamma$. (Section \ref{sec:watSuppCorr})
            \item \textbf{Uncertainty estimation ($\epsilon_i$):} During training and inference, $\epsilon_i \in \mathbb{R}, i=1,...,I$ are sampled from $\mathcal{N}(0,1)$, thereby enabling the generation of diverse yet plausible distributions of network parameterizations via the hypernetwork (Section \ref{sec:uncEst})
        \end{itemize}

    \subsection{Network Architecture}
        The hypernetwork $\mathcal{H}$ takes hyperparameters $\theta$ as input. It contains 4 dense layers separated by 3 ReLU activation layers. The dimensionality of the hidden layers is set to 16, and the output size is identical to the number of weights in the encoder $\mathcal{E}$.

        The encoder $\mathcal{E}$ consists of 5 convolutional blocks and is followed by a dense layer. The output size is equal to the number of parameters provided to the physical model. Each convolutional block contains 4 layers: Two 1-dimensional convolutional layers followed by instance normalization and ReLU activation. The size of the convolutional kernel is set to 5. The feature size of the convolutional layers in the first block is set to 16 and doubled in every subsequent block. A residual connection is implemented at the beginning and added to the output of each block. After every block, a downsampling layer is applied.
        
    \subsection{Physical Model}
        The physical model $\mathcal{P}$ combines the individual spectral components (metabolite basis set $\mu$, baseline $b$, water suppression correction $w$) inside the decoder $\mathcal{D}$. It receives predicted parameters $\mathcal{E}(s)=p$ from the encoder and the basis set $\mu(t)$ containing metabolite and macromolecular signal. The model parameters $p$ include the scalar amplitudes $A$ of the basis set, frequency correction $\delta f$,  signal decay $d$, constant phase $\phi_0$, linear phase $\phi_1$, baseline $b$, and water suppression correction function $w$. 

        The time domain basis functions are linearly combined and after the Fourier transformation the preliminary spectrum $f'$ is obtained.
        \begin{align}
            \mathfrak{d}(t) =&  \text{exp}(- d t) \\
            \mathfrak{f}(t) =&  \text{exp}(-i 2\pi \delta f t) \\
            f'(\nu) =& \text{fftshift}(\text{fft}(\sum_m A_m \mu_m(t) \mathfrak{d}(t) \mathfrak{f}(t)))
        \end{align}
        
         The preliminary spectrum $f'$ is combined with the modeled water suppression correction $w$ and the baseline $b$.
        \begin{align}
            f(\nu) = f'(\nu) w(\nu) + b(\nu)
        \end{align}
        Instead of applying a phase to the fit, the input spectrum $s'$ is phase corrected to align with the basis set.
        \begin{align}
            s(\nu) = s'(\nu) ~\text{exp}(i(\phi_0 + \phi_1 * \nu))
        \end{align}
        The phase corrected spectrum $s$ is as target spectrum during in the loss. 
        
        \subsubsection{Baseline construction} \label{sec:baseline}
            The baseline is modeled with spline functions parametrized by spline node locations and amplitudes. The spline amplitudes are provided by the encoder model $\mathcal{E}$, and the node locations (internode interval) are conditioned by a hyperparameter $\beta$. $\beta$ is a continuous parameter normalized in the range $[0,1]$. It generates equidistant nodes with a spacing of 0.1 to 0.2 ppm.

            The baseline flexibility is dependent on the hyperparameter $\alpha \in [0,1]$, which balances mean-squared error with a term that penalizes baseline $b$ flexibility $\mathcal{L}_{flex}$.
            \begin{align}
                \mathcal{L}(f,s,\alpha,b) = (1-\alpha) MSE(f,s) + \alpha \mathcal{L}_{flex}(b)
            \end{align}
            Here, $\alpha$ balances the relative importance of fitting accuracy and baseline flexibility. $\mathcal{L}_{flex}$ takes the baseline $b$ of the modeled fit $f$ as input, and penalizes high baseline variability.            
            \begin{align}
                \mathcal{L}_{flex}(b) = \sum_i |b[i]-b[i+1]|^2
            \end{align}
            Here $b$ represents the baseline, and $b[i]$ the amplitudes of the interpolated splines. $i$ are the vector indices of the baseline $b$, understood as a 1-dimensional vector.
            
        \subsubsection{Water suppression correction} \label{sec:watSuppCorr}
            The water suppression correction function is controlled by a hyperparameter $\gamma\in\theta$, which is provided as input to the hypernetwork $\mathcal{H}(\theta)$. $\gamma$ only takes binary values 0 or 1. If $\gamma$ is set to 0, the water suppression correction is turned off, thereby setting $w=1$. For $\gamma=1$, a Gaussian correction function is modeled, which is parametrized by a mean $\mu_w$, a standard deviation $\sigma_w$, and an amplitude $A_w$. 
            \begin{align}
                w(\nu) = 
                \begin{cases}
                    \frac{A_w}{\sqrt{2\pi}} * \text{exp}(-\frac{1}{2} \left(\frac{\nu-\mu_w}{\sigma_w}\right) ^2) & \text{if } \gamma == 1 \\
                    1 & \text{if } \gamma == 0
                \end{cases}
            \end{align}
            The water suppression parameter $\mu_w, \sigma_w, A_w$ are predicted by the encoder model $\mathcal{E}$.

    \subsection{Uncertainty estimation} \label{sec:uncEst}
        Our approach extends the uncertainty estimation in PHIVE \cite{shamaei2026phive} by integrating a Bayesian hypernetwork (BHN) \cite{ekmekci2023quantifying} to predict aleatoric and epistemic uncertainty from a single model \cite{chan2024estimating}. 

        To effectively assess uncertainties, we use a Bayesian hypernetwork together with a variational encoder. BHNs receive random noise and stochastically generate weights for the encoder model. In this case, the hypernetwork receives, in addition to the hyperparameters $\alpha, \beta,\gamma\in\theta$, a vector $\epsilon_i \in \mathbb{R}^I, \epsilon_i \in\theta$ for $i=1,...,I$. $\epsilon_i$ is sampled from a normal distribution $\epsilon_i \sim \mathcal{N}(0,1)$. 
        
        The predicted encoder model $\mathcal{E}$ does not predict the metabolite concentration/basis set amplitudes $A_m$ directly; the Variational encoder estimates a Gaussian distribution $\mathcal{N}(\mu_m,\sigma_m)$ for each metabolite. The encoder predicts $\mu_m,\sigma_m$, individual concentrations are derived as $\mu_m + \sigma_m*\eta_m$, where $\eta_m \sim \mathcal{N}(0,1)$. 

        The training process is extended by a loss term $\mathcal{L}_{unc}$ similar to the one presented in \cite{shamaei2026phive}: 
        \begin{align}
            \mathcal{L}(f,s,\alpha,b,\sigma) &= (1-\alpha) MSE(f,s) \\
            &+ \alpha \mathcal{L}_{flex}(b) + \mathcal{L}_{unc}(\sigma) \\
            \mathcal{L}_{unc}(\sigma) &= -\frac{1}{2} \sum_{m=1}^M (1+\text{ln}\sigma_m^2+\sigma_m^2)
        \end{align}
        
        During inference, the BHM approach enables stochastic sampling of an ensemble of encoder models $\mathcal{E}^j, j=1,...,J$. For each model, the hyperparameters $\epsilon^j_i, i=1,..,I$ are randomly sampled from $\mathcal{N}(0,1)$. Each encoder model $\mathcal{E}^j$ predicts for every metabolite a mean $\mu^j_m$ and standard deviation $\sigma^j_m$. The ensemble mean and uncertainty for each metabolite are derived as follows:
        \begin{align}
            \bar{\mu}_m &= \frac{1}{J}\sum_{i=1}^J \mu^j_m \\
            \hat{\sigma}^2_m &= \underbrace{\frac{1}{J}\sum_{i=1}^J (\sigma^j_m)^2}_\text{AU} + 
            \underbrace{\left(\frac{1}{J}\sum_{i=1}^J (\mu^j_m)^2\right) -  \bar{\mu}^2}_\text{EU}
        \end{align}
        The first term represents the aleatoric uncertainty (AU) and the second term the epistemic uncertainty (EU). In this study, following the recommendation in \cite{lakshminarayanan2017simple} and implementation in \cite{shamaei2026phive}, $J=5$ encoder models were sampled. Furthermore, in line with \cite{chan2024estimating}, the vector-size of $\epsilon_i \in \mathbb{R}^I$ is set to $I=8$.

    \subsection{Training details}
        MRSI spectra were pre-processed by normalizing the maximum magnitude to 1 and cropping spectra to the range between 4.2 and 1 ppm. During training, data augmentation was performed using random scaling $\pm10\%$, random constant phase (0-2$\pi$), linear phase (-$\pi$,$\pi$), frequency shift ($\pm$50 Hz) and random decay (0-0.2$s^{-1}$). Due to dwell time difference during the acquisition, separate networks are trained for 3T and 7T MRSI.
        
        The networks were trained for 1000 epochs, using the Adam optimizer with a learning rate of 0.001, which was quartered every 400 epochs. The exponential decay rates of the first and second momentum of Adam $\beta_1$ \& $\beta_2$ were set to 0.9 and 0.999. The network was implemented using PyTorch 2.0.1 and CUDA 11.7 packages in Python 3.8.13. The model training was performed on a PowerEdge R7525 server (Dell) with 64 CPU cores (AMD EPYC7542 2.90GHz, 128M Cache, DDR4-3200), 512 GB CPU RAM (RDIMM, 3200MT/s), 3 GPU NVIDIA Ampere A40 (PCIe, 48GB GPU RAM) running Rocky Linux release 8.8 (Green Obsidian).
    
    \subsection{Data acquisition}
        In-vivo MRSI data were acquired with 3D $^1$H-FID ECCENTRIC pulse sequence using a 7T MR scanner (MAGNETOM Terra, Siemens Healthineers, Erlangen, Germany) equipped with 1Tx/32Rx head coil (NovaMedical, Wilmington, MA, USA), and 3T MR scanner (MAGNETOM Prisma, Siemens Healthineers, Erlangen, Germany) equipped with a 32-channels receive array. 3D $^1$H-FID ECCENTRIC was acquired on 9 subjects at 7T and 12 subjects at 3T.  
        
        The datasets were acquired at 7T with a  FoV of $220 \times 220 \times 105 \text{mm}^3$ and matrix size of $22 \times 22 \times 10$ (10mm isotropic), $64 \times 64 \times 31$ (3.3mm isotropic) and $110 \times 110 \times 51$ (2mm isotropic), and spectral bandwidth of 2280 Hz, the number of acquired FID points was set to 453 (10mm, 3.4mm) or 410 (2mm). Echo time (TE) of 0.9 ms and repetition time (TR) of 275ms (10mm, 3.4mm) or 188ms (2mm) was used. 

        The datasets acquired at 3T had a matrix size of $64 \times 64 \times 31$, a FoV of $220 \times 220 \times 105 \text{mm}^3$ FoV, a spectral bandwidth of 1260 Hz, and 453 FID points. For the sequence, a TE of 0.9 ms and a TR of 350 ms were used. 
    
    \subsection{Processing pipeline}
        The acquired ECCENTRIC k-space data were processed with our end-to-end deep learning MRSI processing pipeline \cite{weiser2025deep} which includes several steps: 1.) non-uniform fast Fourier transform, 2.) ESPIRIT coil combination \cite{uecker2014espirit}, 3.) B0 correction, 4.) water and lipid removal with WALINET \cite{weiser2025walinet} and WALINET+ \cite{weiser2025walinet, weiserwalinet+} 5.) low-rank model \cite{klauser2024eccentric} denoising, 6.) DeepER image reconstruction \cite{weiser2025deep}. Spectral fitting was performed with the proposed HyperFitS and conventional LCModel \cite{provencher1993estimation} as the gold standard method.

        Additional pre-processing included the backward prediction of the first 2 missing time points during echo time using auto-regressive modeling. A qsine filter was applied in the time domain, and MRSI data acquired with 410 time points were zero-padded to 453 time points.

    \subsection{Monte Carlo analysis}
        To investigate the ability of HyperFitS to separate the spectral overlap of individual metabolites we calculated the correlation coefficients between the metabolite concentrations predicted by HyperFitS using a representative collection of 7T and 3T in-vivo spectra. 500 levels of random Gaussian noise were generated and added to the collection of spectra. The Gaussian noise was sampled with a mean of 0 and a standard deviation of $\frac{1}{50}$, similar to \cite{shamaei2023physics}. The spectra amplified by random noise were fitted by the dedicated 3T and 7T networks. Predicted concentrations of each metabolite were retrieved. For each spectrum, the Pearson correlation coefficients were computed between all predicted metabolite concentrations.

\section{Results}
    \begin{figure*}[ht!]
		\centering
		\includegraphics[width=1.0\linewidth]{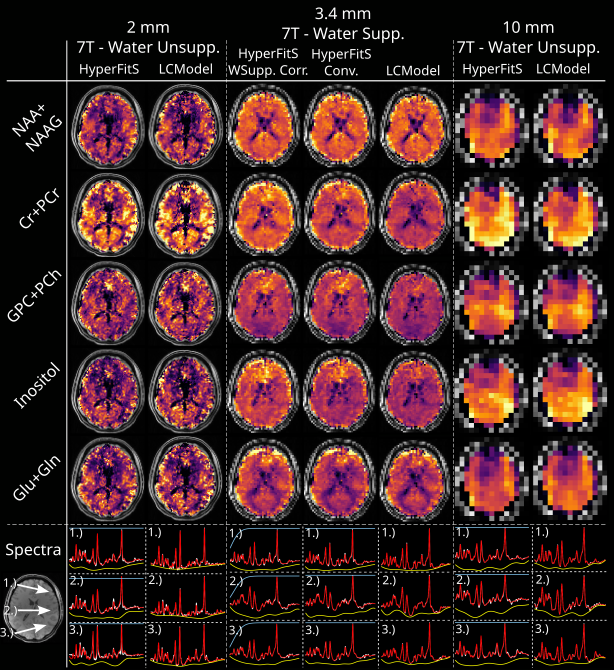}
		\caption{A qualitative comparison showing metabolic maps of NAA+NAAG, Cr+PCr, GPC+PCh, inositol, and Glu+Gln, at 7T quantified with HyperFitS and LCModel. Individual representative spectra are displayed below. Processing times for a single MRSI volume of HyperFitS and LCModel are 10sec vs $\sim$1h for 3.4mm, and 90sec vs $\sim$8h for 2mm}
        \label{fig:qualitative7T}
	\end{figure*}
    
    First, figure \ref{fig:qualitative7T} presents a qualitative comparison of HyperFitS and LCModel at 7T. Metabolite maps are presented for MRSI acquired at 7T at 2 mm, 3.4 mm and 10 mm isotropic resolution. Additionally, the effects on metabolite maps of water suppression correction. 

    LCModel and HyperFitS show a good agreement of spatial distribution across several metabolite maps. Similar overall metabolite concentration across the brain is visible at 7T at 3.4 mm. In particular, NAA+NAAG, GPC+PCh or Glu+Gln show good agreement. Clearly visible anatomical features, such as the extension of the lateral ventricles in NAA+NAAG and Glu+Gln are well correlated in both methods. Additionally, comparison of results obtained with and without correction of water suppression effects are shown for data acquired at 3.4 mm resolution. The modeled water suppression correction function is displayed in blue, overlaid with sample spectra at the bottom of figure \ref{fig:qualitative7T}. The water suppression correction effect is clearly visible, especially for metabolite such as Cr+PCr and Inositol that have peaks located near the resonance frequency of water at 4.7 ppm. The water suppression reduces the signal of Cr+PCr and Inositol peaks at 3.9 ppm, and correction of this effect results in higher concentration estimated for these metabolites.

    \begin{figure*}[ht!]
		\centering
		\includegraphics[width=1.0\linewidth]{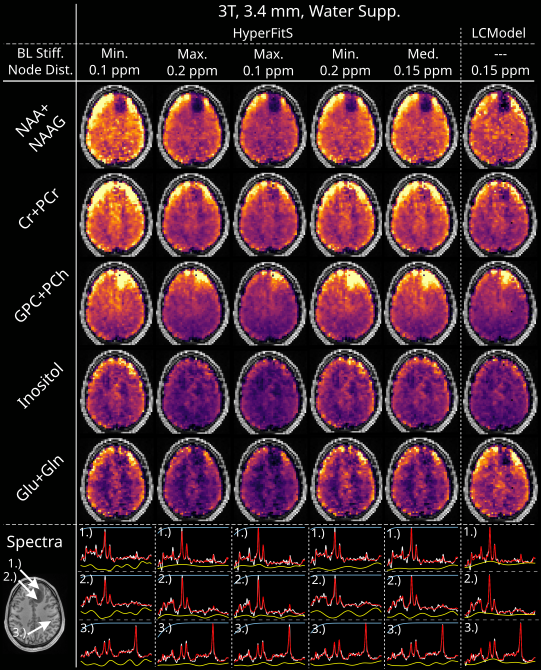}
		\caption{Metabolic maps of NAA+NAAG, Cr+PCr, GPC+PCh, Inositol, and Glu+Gln quantified with HyperFitS and LCModel from 3D MRSI acquired at 3T in a glioma patient. Maps are compared for different baseline modeling. Individual representative spectra are displayed below.}
        \label{fig:qualitative3T}
	\end{figure*}
    
    Figure \ref{fig:qualitative3T} presents the effects of  baseline parametrization at 3T, by varying baseline flexibility and node distance. Metabolic maps show higher concentration estimates and noise with the flexible baseline. Furthermore, a very flexible baseline can contribute to reduced contrast between tumor and healthy brain in the glioma patient.

    \begin{figure*}[ht!]
		\centering
		\includegraphics[width=\linewidth]{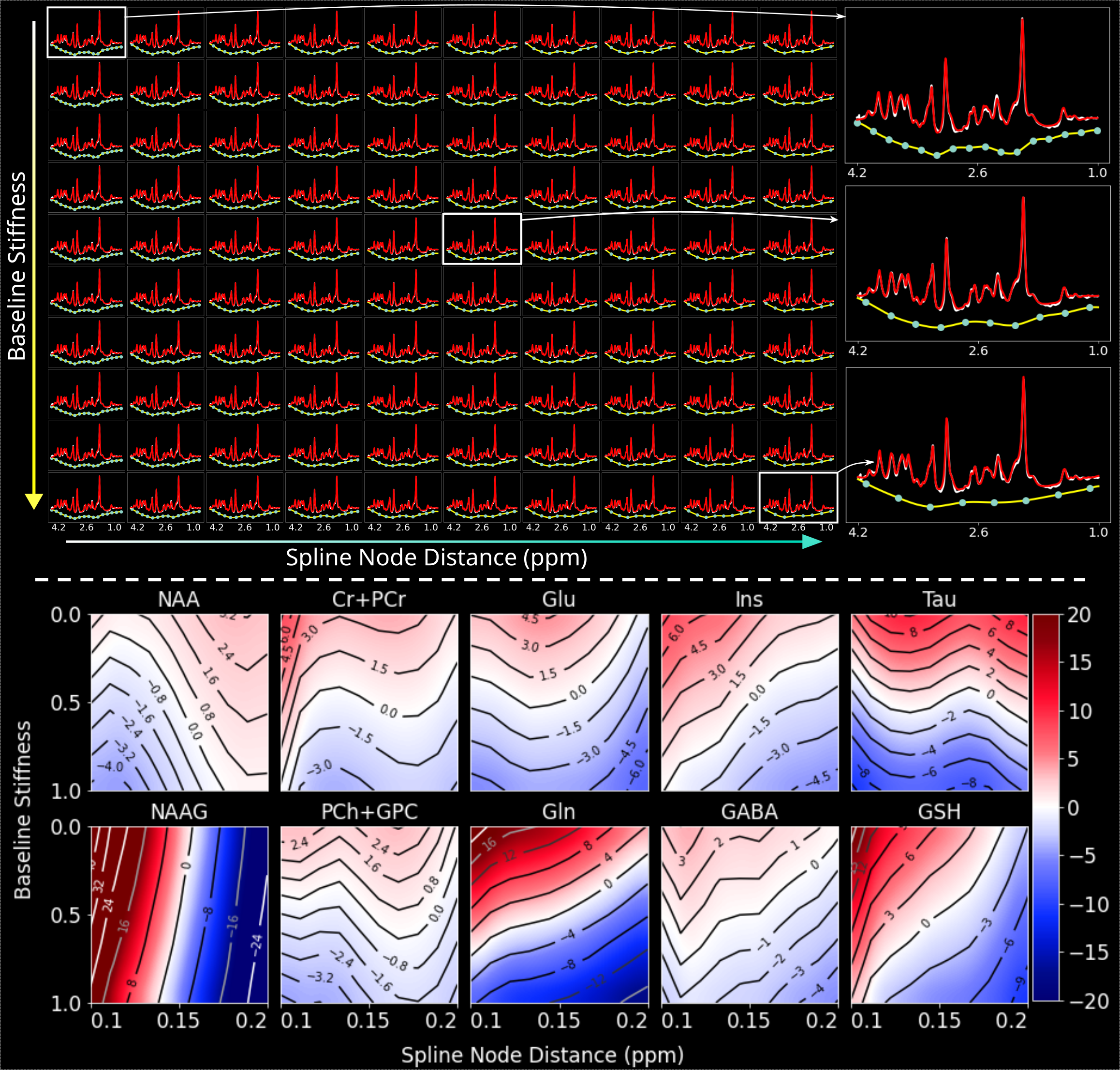}
		\caption{\textbf{Top:} A representative spectrum fitted by varying the hyperparameters for baseline flexibility (vertical) and spline node distance (horizontal) in increments of $\frac{1}{10}$ of the total range. Three cases are shown enlarged on the right. The original spectrum is white, fit is red, baseline is yellow, and spline nodes are blue. \textbf{Bottom:} Contour plots of metabolite concentrations. A subject acquired at 7T with 3.4mm resolution was evaluated for 100 different baseline configurations. Baseline flexibility and spline node distance were changed over the hyperparameter range with increments of $\frac{1}{10}$ of the total range. Values show the difference to the center value (0.5 flexibility, 0.15 ppm node distance) in percent.}
        \label{fig:baseline}
	\end{figure*}
    
    Figure \ref{fig:baseline} shows a more detailed investigation of the effects of changing baseline flexibility and spline node distance. In the upper panel a representative in-vivo 7T MRSI spectrum has been evaluated with 100 hypernetwork realizations, varying baseline flexibility from 0 to 1 and node distance from 0.1 to 0.2 ppm in increments of $\frac{1}{10}$ of the total interval. In the bottom panel the change in metabolite concentration are displayed in percent of the central value (0.5 flexibility, 0.15 ppm node distance). It can be seen that metabolites that have singlet peaks or larger signals such as NAA, Cr+PCr, PCh+GPC and Glu the variability in estimated concentrations due to baseline parametrization is $\pm 4\%$. On the other hand, the baseline parametrization increases the variability of estimated concentration up to $\pm 25\%$ for  metabolites that have lower signals such as NAAG, Gln, Tau and GSH. Across most metabolites, an increased baseline flexibility is correlated with increased metabolite concentrations. Notable metabolites that reflect this phenomenon are Gln and Tau. The concentration of NAA and NAAG changes in opposite directions with respect to the spline node distance: NAA is more likely to increase for larger node distance, while NAAG decreases.

    \begin{figure*}[ht!]
		\centering
		\includegraphics[width=1.0\linewidth]{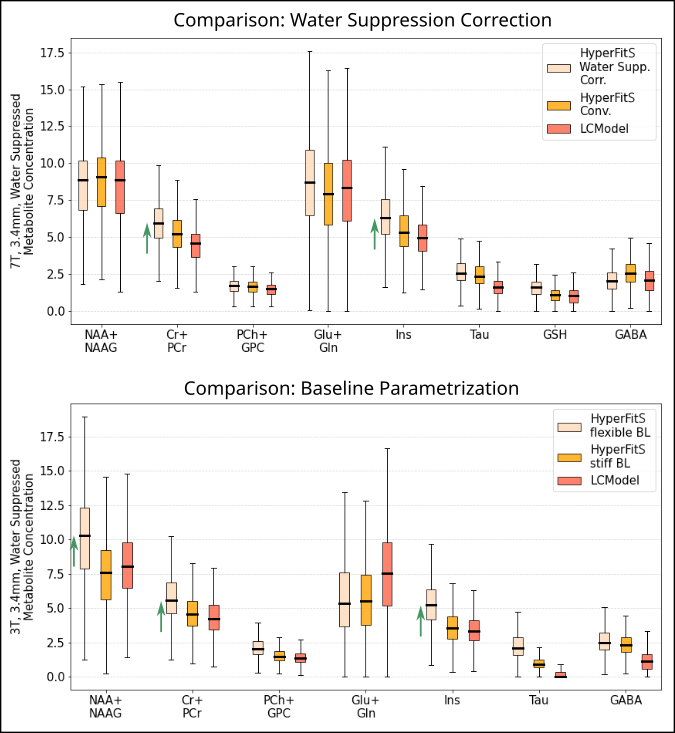}
		\caption{Metabolite concentration boxplots for metabolites NAA+NAAG, Cr+PCr, PCh+GPC, Glu+Gln, Ins, Tau, GSH, and GABA. Evaluations are performed for 7T (top) and 3T (bottom). The 7T was evaluated with water suppression correction by HyperFitS. The 3T was processed with different baseline configurations by HyperFitS. Comparison to LCModel is included.}
        \label{fig:boxplot}
	\end{figure*}

    Figure \ref{fig:boxplot} shows boxplot comparison between HyperFitS and LCModel for metabolite concentrations at 3T and a 7T. At 7T the HyperFitS results are shown with and without water suppression correction. At 3T HyperFitS results are shown for stiff and flexible baseline. In general close agreement is noted between HyperFitS and LCModel, and in particular it can be seen that: 1) correction of water suppression by HyperFitS provides higher concentrations for Cr+PCr and Ins, 2) more stiff baseline provides better agreement of HyperFitS with LCModel.

    \begin{figure*}[ht!]
		\centering
		\includegraphics[width=1.0\linewidth]{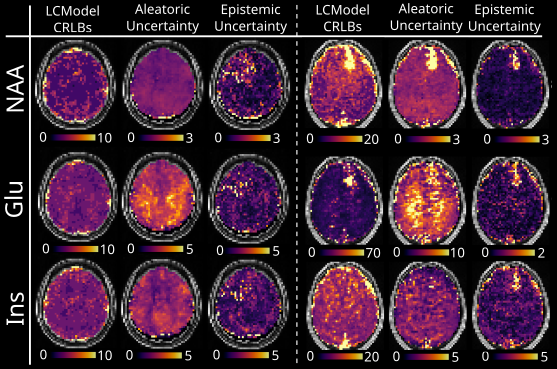}
		\caption{Uncertainty maps for the metabolites NAA, Glu and Ins at 7T (left) and 3T (right). LCModel uncertainties are computed as CRLBs, HyperFitS uncertainties are shown with Aleatoric and Epistemic uncertainty.}
        \label{fig:uncertainty}
	\end{figure*}

    Figure  \ref{fig:uncertainty} shows a comparison between LCModel generated Cramer-Rao lower bounds (CRLB) and aleatoric (AU) \& epistemic uncertainty (EU) maps from HyperFitS. 
    The uncertainty values lie in different numerical ranges, with CRLBs ranging form 0 to 20, and in rare cases exceeding 70. AU and EU generally lie between 0 and 5. In particular, AU values increase in brain areas where metabolite concentrations such as glutamate in white matter or NAA in glioma tumor. EU reveals a similar pattern with larger uncertainty in areas with low metabolite levels.
    
    \begin{figure*}[t]
		\centering
		\includegraphics[width=1.0\linewidth]{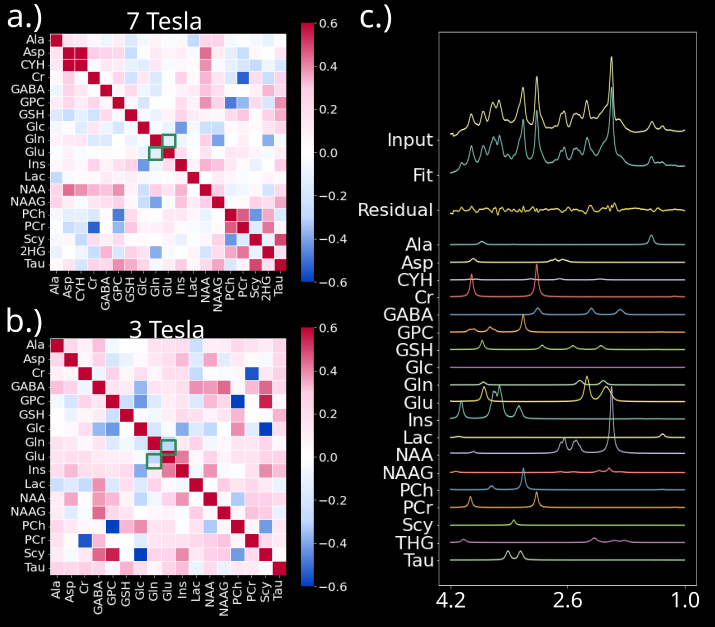}
		\caption{Correlation matrix of in-vivo metabolite concentrations fitted by HyperFitS at a.) 7T and b.) 3T. c.) A spectrum acquired at 7T with its fit and individual metabolite components.}
        \label{fig:correlation}
	\end{figure*}

    Figure \ref{fig:correlation} displays the pair-wise correlation coefficient matrices between metabolite concentrations fitted by HyperFitS from 7T and 3T in-vivo MRSI data. The correlation matrices for the 7T and 3T spectra reveal a strong anti-correlation between Cr \& PCr and GPC \& PCho. Furthermore, the correlation matrix at 7T shows smaller anti-correlations between multiple metabolite pairs, compared to 3T, which is consistent with less spectral overlap and larger peak separation at 7 Tesla. The full count of significant correlations at each field strengths is provided in Supplementary Table \ref{tab:table_corr}, showing that the number of significant correlations at 7T decreases compared to 3T. In particular, correlation coefficients of Glu \& Gln show increased anti-correlation at 3T than 7T, pointing to better separation of these important metabolites at 7T. The effect of different baseline parametrization on the correlation between metabolite concentrations is provided the Supplementary Figures \ref{fig:correlation2_3T} and \ref{fig:correlation2_7T}. In particular, it can be seen that flexible baselines tend to promote positive correlations while stiffer baselines negative correlations.

\section{Discussion}
	We demonstrate HyperFitS, a fast and accurate hypernetwork to fit spectra for the metabolic quantification of 3-dimensional whole-brain MRSI. HyperFitS eliminates time-consuming iterative fitting by utilizing a GPU-accelerated deep learning model. HyperFitS provides effective spectral quantification, which allows for rich customizability during inference, such as enabling water suppression correction or baseline tuning that can adapt to a multitude of protocols, magnetic fields and data quality. Thereby, HyperFitS can simplify time-consuming spectral quantification and analysis.

    Considering that in recent years the resolution and matrix sizes of MRSI acquisitions have continuously increased, there is a requirement for fast and effective processing modules that can handle increasingly large datasets. Compared to conventional spectral quantification tools, such as LCModel, which can require 1 hour to process a single 3.4 mm isotropic MRSI volume and 8 hours for 2 mm MRSI, HyperFitS can significantly reduce the computational workload and quantify an MRSI volume in less than 10 seconds. 

    Importantly, the fitting accuracy is maintained for fast processing time and HyperFitS showed good agreement with gold-standard LCModel quantification providing similar mean values and confidence intervals for multiple metabolic maps at different field strengths and acquisition protocols. 
    
    The baseline modeling in HyperFitS uses low-degree polynomial functions similar to previous methods \cite{provencher1993estimation} \cite{shamaei2026phive}. However, due to extensive processing times of conventional methods or the need to retrain deep-learning models, these methods fix the number of baseline nodes adjusting only the amplitude of polynomial functions. In contrast, HyperFitS allows independent adjustment of the two parameters controlling the baseline flexibility and node distance during inference. This enhances the ability to model complex baselines and allows to adapt to different data quality including large baseline distortions caused by technical imperfections such as B0 inhomogeneity, poor water suppression, or lipid contamination.
    
    Water suppression pulses can reduce the amplitude of metabolite peaks that are close to the water signal. Previous deep learning models do not correct for this effect, therefore providing potentially inaccurate metabolite quantification. To account for the water suppression effects in the quantification of metabolite concentrations we introduced in the physical model of HyperFitS an inverted Gaussian function that is learned as a hyperparamter. This approach provides increased concentration and gray-white matter contrast for metabolites such as Cr+PCr and Inositol that are more susceptible to water suppression.

    In contrast to previous studies that focus on single field MRS data acquired with a single protocol, this work extends the application of deep-learning fitting to multiple fields and multiple acquisition protocols, comparing results from 3T and 7T MRSI. In particular, analysis of correlations coefficients between metabolite quantification shows lower correlation between glutamate and glutamate at 7T, consistent with larger spectral separation and less peak overlap compared to 3T. Similar effect can be noted for correlations of Cr to PCr, and PCh to GPC. 

    The main limitation of HyperFitS is related to the fact that it was trained only on short-echo 1H-MRSI data. In its current implementation the use of HyperFitS for fitting longer spin-echo MRSI or other nuclei would require retraining the network with corresponding data. However, because of the physics-based approach the amount of data required for training is low and enough spectra can be generated from high-resolution MRSI in ten subjects.

\section{Conclusion}
    Fast and accurate metabolic quantification is a key requirement for the applicability of 3-dimensional whole-brain $^1$H MRSI. Conventional spectral fitting methods can require long runtimes for high-resolution 3D MRSI. In recent years, deep learning methods have gained increasing popularity due to their speed to quantify MRSI. However, these methods are often inflexible and require inefficient retraining to support new, unseen protocols. 
   
    In this work, we introduce a hypernetwork-based framework that can adapt to the quantification of MRSI data acquired at multiple fields and protocols. The proposed model enables integrated water suppression correction and a continuously adjustable baseline, allowing seamless adaptation to different parametrizations without retraining. This flexibility, combined with rapid inference, represents a substantial step toward robust and scalable MRSI quantification. We expect that this methodology has the potential to facilitate the clinical application and integration of MRSI into the radiological workflow.

\vspace{1cm}
\noindent\textbf{Acknowledgements: }This work was supported by FWF research grants P34198, National Institute of Health research grants 2R01CA211080-06A1, R01CA255479, P50CA165962, P41EB015896, R00HD101553, and the Athinoula A. Martinos Center for Biomedical Imaging

\noindent\textbf{Ethical Standards: } The work follows appropriate ethical standards in conducting research and writing the manuscript, following all applicable laws and regulations regarding treatment of animals or human subjects.

\noindent\textbf{Conflict of Interest: } Antoine Klauser is currently employed by Siemens Healthineers.

\noindent\textbf{Data availability: } Testing data can be obtained from the authors based on reasonable request and institutional approved data sharing agreement.

\bibliographystyle{unsrt}
\bibliography{bibliography}
\newpage
\clearpage

\section*{Supplementary Material}
\setcounter{figure}{0}
\setcounter{table}{0}
\renewcommand{\figurename}{Supplementary Figure} 
\renewcommand{\tablename}{Supplementary Table} 

\begin{table*}[h]
    \centering
    \begin{tabular}{|ccc||ccc|}
        \hline
        \multicolumn{3}{|c||}{7 Tesla} & \multicolumn{3}{c|}{3 Tesla} \\
        \hline
        15.0/4.0 & 10.0/3.0 & 7.0/4.0 & 24.0/4.0 & 24.0/3.0 & 19.0/4.0 \\
        16.0/4.0 & 8.0/4.0 & 8.0/4.0 & 18.0/4.0 & 16.0/7.0 & 15.0/4.0 \\
        13.0/6.0 & 8.0/4.0 & 10.0/4.0 & 7.0/6.0 & 9.0/6.0 & 12.0/4.0 \\
        \hline
    \end{tabular}
    \caption{Percent of significantly (more than 0.3 or less than -0.3) correlated or anti-correlated metabolites for increasing baseline flexibility and node distance. Baseline flexibility parameters from top to bottom: 0.0, 0.5, 1.0. Node distance from left to right: 0.1 ppm, 0.15 ppm, 0.2 ppm. 7T left, 3T right. The first value shows the correlated percentage, the second shows the anti-correlation percentage.}
    \label{tab:table_corr}
\end{table*}

\begin{figure*}[t]
    \centering
    \includegraphics[width=1.0\linewidth]{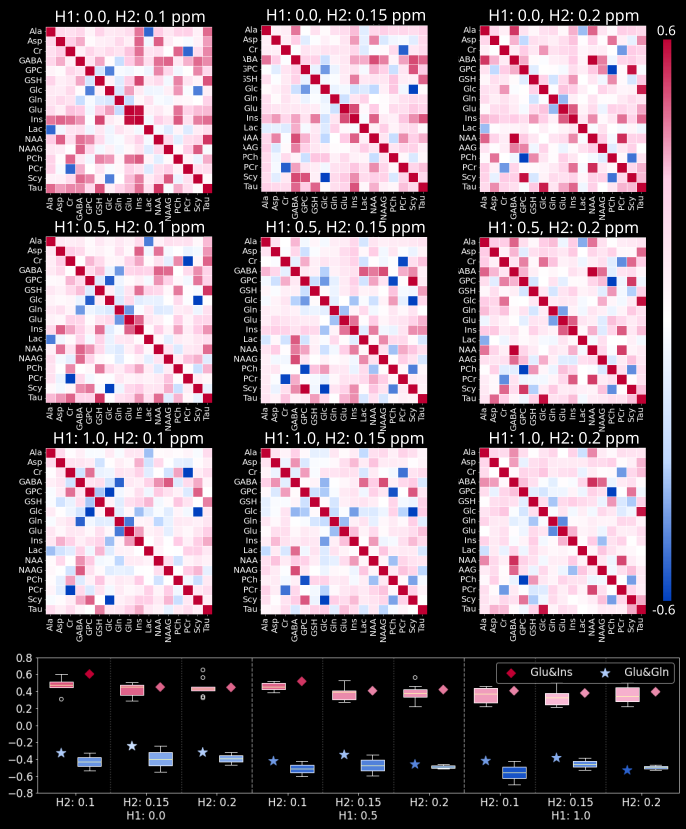}
    \caption{Top: Correlation matrix of Monte Carlo analysis for subjects acquired at 3T. Each subject was evaluated for 9 different baseline configurations. Bottom: Boxplots showing correlation values of metabolite pairs that are less than -0.2 or larger than 0.2 in all baseline configurations. Glu \& Ins and Glu \& Gln are plotted separately as diamond/star. The color selection is based on the median values.}
    \label{fig:correlation2_3T}
\end{figure*}

\begin{figure*}[t]
    \centering
    \includegraphics[width=1.0\linewidth]{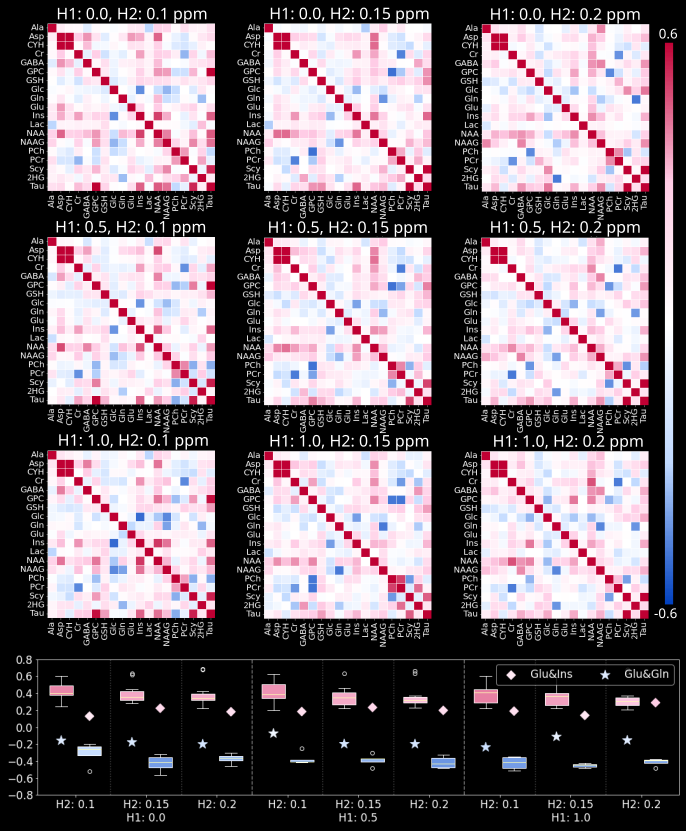}
    \caption{Top: Correlation matrix of Monte Carlo analysis for subjects acquired at 7T. Each subject was evaluated for 9 different baseline configurations. Bottom: Boxplots showing correlation values of metabolite pairs that are less than -0.2 or larger than 0.2 in all baseline configurations. Glu \& Ins and Glu \& Gln are plotted separately as diamond/star. The color selection is based on the median values.}
    \label{fig:correlation2_7T}
\end{figure*}


\end{document}